\documentclass[12pt,logo]{spell}
\usepackage{nicefrac}           

\usepackage{multirow}
\usepackage{array}
\usepackage{adjustbox}
\usepackage{makecell}
\usepackage{longtable}
\usepackage{arydshln}           

\usepackage{wrapfig}
\usepackage{subcaption}
\usepackage{rotating}
\usepackage[abs]{overpic}
\usepackage{tikz}               

\usepackage[most]{tcolorbox}

\usepackage{listings}
\usepackage{csquotes}

\usepackage{fontawesome}

\usepackage[capitalize]{cleveref}

\usepackage{tocloft}

\usepackage{lipsum}
\usepackage{float}

\usepackage{siunitx}
\usepackage{adjustbox}
\usepackage[table]{xcolor}
\usepackage{graphicx}
\usepackage{subcaption}

\usepackage{multirow}
\usepackage{pifont}

\definecolor{bluelink}{RGB}{0,113,188}
\definecolor{greenlink}{RGB}{0,188,113}
\definecolor{linkteal}{RGB}{0,102,102}
\definecolor{linkblue}{RGB}{0,51,102}
\hypersetup{
    colorlinks=true,
    citecolor=green!93!black,
    linkcolor=linkblue,
    urlcolor=bluelink
}

\definecolor{codekeyword}{rgb}{0.0, 0.0, 0.5}
\definecolor{codecomment}{rgb}{0.0, 0.5, 0.0}
\definecolor{codestring}{rgb}{0.56, 0.0, 1.0}

\lstdefinestyle{pythonstyle}{
    language=Python,
    basicstyle=\ttfamily\small,
    keywordstyle=\color{codekeyword}\bfseries,
    commentstyle=\color{codecomment}\itshape,
    stringstyle=\color{codestring},
    showstringspaces=false,
    breaklines=true,
    tabsize=4,
    numbers=none,
    frame=none,
    backgroundcolor=\color{white},
    captionpos=b,
    morekeywords={self, __init__, __name__, __main__},
}
\def\dataset{{\textit{SceneOnto}}}
\def\evaluator{{\textit{SceneCritic}}}

\newcommand{\authorssep}{~~~~~~~~}

\title{\center SceneCritic: A Symbolic Evaluator for 3D Indoor Scene Synthesis}

\author{Kathakoli Sengupta \authorssep{} Kai Ao \authorssep{} Paola Cascante-Bonilla\\
    \vspace{-12pt}
    {\normalfont\small\texttt{\{ksengupta, paola\}@cs.stonybrook.edu}}\\
    \vspace{8pt}
    Stony Brook University\\
    \vspace{8pt}
    {\small
    \faGlobe\ \href{https://lab-spell.github.io/SceneCritic}{Project Page} \quad
    \faGithub\ \href{https://github.com/DIASENGUPTA/SceneCritic}{Repository}
    }
}

\begin{abstract}
Large Language Models (LLMs) and Vision-Language Models (VLMs) increasingly generate indoor scenes through intermediate structures such as layouts and scene graphs, yet evaluation still relies on LLM or VLM judges that score rendered views, making judgments sensitive to viewpoint, prompt phrasing, and hallucination.
When the evaluator is unstable, it becomes difficult to determine whether a model has produced a spatially plausible scene or whether the output score reflects the choice of viewpoint, rendering, or prompt.
We introduce \textbf{\evaluator{}}, a symbolic evaluator for floor-plan-level layouts. 
\evaluator{}'s constraints are grounded in \textbf{\dataset{}}, a structured spatial ontology we construct by aggregating indoor scene priors from 3D-FRONT, ScanNet, and Visual Genome. \evaluator{} traverses this ontology to jointly verify semantic, orientation, and geometric coherence across object relationships, providing object-level and relationship-level assessments that identify specific violations and successful placements.
Furthermore, we pair \evaluator{} with an iterative refinement test bed that probes how models build and revise spatial structure under different critic modalities: a rule-based critic using collision constraints as feedback, an LLM critic operating on the layout as text, and a VLM critic operating on rendered observations.
Through extensive experiments, we show that \textit{(a)} \evaluator{} aligns substantially better with human judgments than VLM-based evaluators, \textit{(b)} text-only LLMs can outperform VLMs on semantic layout quality, and \textit{(c)} image-based VLM refinement is the most effective critic modality for semantic and orientation correction.
\end{abstract}

\begin{document}

\maketitle

\section{Introduction}
\label{sec:intro}

The generation of 3D indoor environments has become central to a range of applications, from training embodied agents that must navigate and manipulate objects in realistic spaces~\cite{xia2026sage, ai2thor, procthor, savva2019habitat}, to virtual reality and robotics simulation. Recent work has demonstrated that Large Language Models (LLMs) and Vision-Language Models (VLMs) can serve as powerful priors for this task, leveraging their world knowledge to produce object layouts that are both diverse and semantically meaningful~\cite{feng2023layoutgpt, yang2024holodeck, bian2025holodeck, sun2025layoutvlm}. These approaches generate scenes through explicit structured representations (e.g., floor plans, scene graphs, spatial constraints), often coupled with refinement strategies and constraint-based optimization to improve physical and semantic plausibility before asset placement and rendering. Implicitly, this pipeline frames scene generation as a test of whether a model can construct and maintain a coherent spatial representation that remains consistent across successive placement and refinement steps.

\begin{figure}[t!]
    \centering
    \includegraphics[width=1\linewidth]{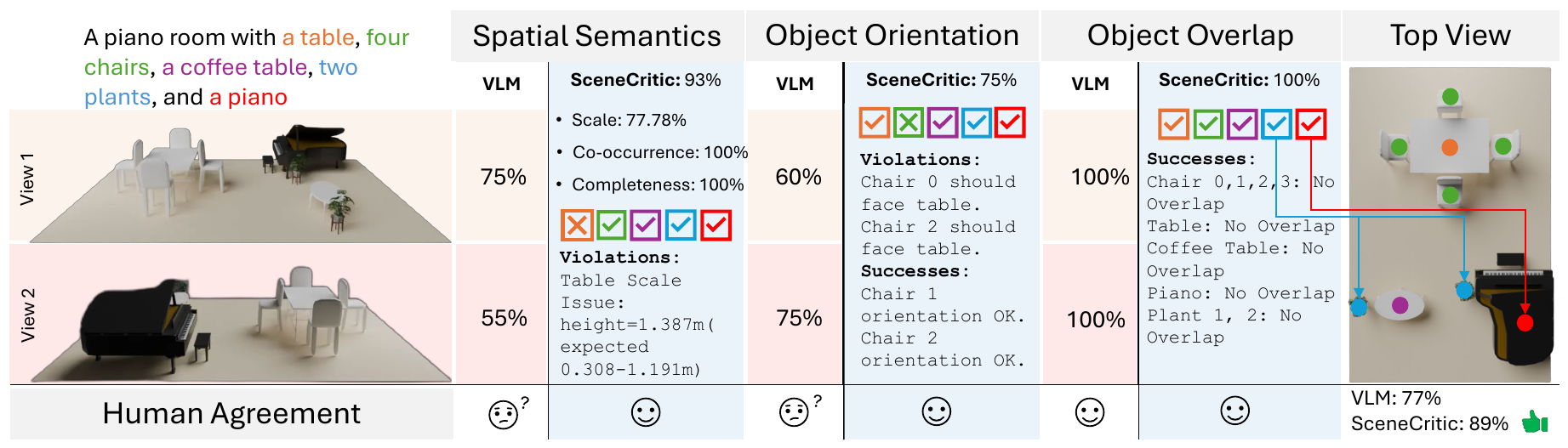}
    \vspace{-0.7cm}
    \caption{\textbf{VLM Evaluation vs. \evaluator{}.} Given a piano room with nine objects, the VLM assigns different scores to the same scene depending on which view is rendered, while \evaluator{} evaluates the layout directly by traversing relational constraints from \dataset{} (e.g., incorrect table scales, chairs not facing the table they co-occur with). This produces stable, object-level assessments that identify specific violations alongside successful placements. \evaluator{} scores also align more closely with human preferences.}
    \vspace{-0.3cm}
    \label{fig:figure1}
\end{figure}

Despite this progress, the evaluation of generated scenes remains surprisingly underexplored. The most common protocol is to ask a VLM to judge a small number of rendered views, but such evaluators are unstable: they miss object relations due to viewpoint and occlusion, are prone to hallucination, and are highly sensitive to prompt design, where minor rephrasing can substantially alter scores~\cite{li2024topviewrs}. 
As \autoref{fig:figure1} illustrates, a VLM evaluator assigns scores of 75\% and 55\% to the same scene depending on the rendered view, highlighting the need for view-independent evaluation.
Several works acknowledge these limitations and resort to human user studies~\cite{ling2025scenethesis, ran2025direct, zheng2025constructing, liu2025agentic}, which are more reliable but expensive to scale and still produce only single scores or coarse rankings without identifying which spatial constraints were satisfied or violated.

As the field of 3D scene generation continues to grow rapidly, especially in embodied and interactive settings~\cite{yang2024physcene, ling2025scenethesis, chen2026scenefoundry, wang2025embodiedgen}, there is a pressing need for a trustworthy and comprehensive evaluation framework. However, designing such an evaluator is far from straightforward, as spatial reasoning involves multiple interrelated concepts including semantic compatibility, geometric property constraints, and inter-object relationships~\cite{chen2024spatialvlm}. Existing programmatic metrics have addressed specific dimensions of this problem but remain narrow in scope. Distributional statistics such as FID, KL divergence over furniture categories, and out-of-bound rates~\cite{feng2023layoutgpt} capture global similarity but cannot diagnose individual placements. Physics-oriented constraints such as collision avoidance, room-layout IoU, and reachability~\cite{yang2024physcene} measure geometric feasibility while leaving semantic coherence unaddressed. Other systems combine collision-free and in-boundary scores with LLM ratings~\cite{chen2025autolayout, yang2025optiscenellmdrivenindoorscene}, but still outsource semantic judgment to a model-based judge. Among existing frameworks, SceneEval~\cite{tam2025sceneeval} goes furthest by combining text-scene fidelity metrics with physics-based plausibility checks, but does not evaluate semantic layout coherence such as whether object co-occurrences, scales, and orientations reflect plausible indoor configurations. Across this landscape, semantic layout plausibility is consistently either delegated to LLM and VLM judges, or completely unaddressed.

In this paper, we introduce \evaluator{}, a symbolic evaluator for indoor floor-plan layouts. \evaluator{}'s constraints are grounded in \dataset{}, a structured spatial ontology dataset we construct by aggregating priors from 3D-FRONT, ScanNet, and Visual Genome, encoding object co-occurrence statistics, expected scales, and canonical orientations. While prior work has encoded indoor spatial knowledge implicitly through learned embeddings~\cite{zhou2019scenegraphnet}, per-scene graph construction~\cite{gu2024conceptgraphs}, or hand-coded procedural rules~\cite{procthor}, these representations are designed for generation or perception, not evaluation. \dataset{} instead constructs explicit relational graphs of conditional co-occurrence probabilities, scale distributions, and canonical orientations per room type, aggregated across multiple datasets. This structure is what enables \evaluator{} to traverse object relationships during evaluation and localize violations to specific object pairs. From these priors, \evaluator{} derives interpretable constraints that score layouts along three axes: semantic coherence, orientation correctness, and overlap. Because all scene generation pipelines produce a layout before rendering, \evaluator{} targets this shared intermediate representation directly. By evaluating at the object and relationship level, it explicitly identifies which objects succeed or violate which constraints.

A stable symbolic evaluator also enables a deeper question: \textit{how do models with different post-training objectives build and revise spatial structure over successive refinement steps?} We pair \evaluator{} with an iterative refinement testbed in which a model repeatedly improves an initial layout under one of several critic modalities: \textit{(a)} a rule-based critic using collision constraints as feedback, \textit{(b)} an LLM critic operating on the layout as text, or \textit{(c)} a VLM critic operating on rendered observations. Because \evaluator{} scores every intermediate layout, the test bed produces refinement trajectories, making it possible to study which errors different models can correct, where they plateau, and whether fixing one violation introduces new ones elsewhere.

Our key contributions are: \textit{(i)} we construct \dataset{}, a dataset-grounded ontology for indoor layout evaluation, including object dimensions, co-occurrence statistics, support surfaces, and orientation preferences; \textit{(ii)} we develop \evaluator{}, a symbolic evaluator for indoor floor-plan layouts with interpretable constraints over semantic coherence, orientation correctness, and overlap, and validate it against human judgment, showing it aligns substantially better than VLM-based evaluators; \textit{(iii)} we introduce an iterative refinement testbed using heuristic, LLM, and VLM critics to probe how models trained under different post-training objectives build, maintain, and revise spatial structure over multiple correction steps; \textit{(iv)} we use \evaluator{} to identify comparative strengths and failure modes across critics and post-training regimes, showing that some text-only LLMs outperform VLMs on semantic layout quality, and that image-based VLM refinement is the most effective critic modality.
\section{Related Works}
\label{sec:related_work}

\textbf{LLM and VLM-Driven 3D Scene Generation.}
Recent work has increasingly adopted LLMs and VLMs for 3D indoor scene generation, leveraging their spatial reasoning and world knowledge to produce structured layouts from text descriptions. LayoutGPT~\cite{feng2023layoutgpt} uses in-context learning to directly predict numerical layouts,
SceneCraft~\cite{hu2024scenecraft} generates Blender code via scene graphs with VLM-based iterative refinement, and I-Design~\cite{ccelen2024design} leverages multi-agent LLM collaboration with scene graph mechanisms. Holodeck~\cite{yang2024holodeck} uses GPT-4 for fully automated environment generation with constraint-based placement, while LLplace~\cite{Yang2024LLplace} and FlairGPT~\cite{Littlefair2025FlairGPT} refine LLM-driven layout generation through supervised fine-tuning and detailed object descriptions, respectively. More recently, LayoutVLM~\cite{sun2025layoutvlm} combines VLM semantic knowledge with differentiable optimization for physically plausible layouts, and Holodeck 2.0~\cite{bian2025holodeck} introduces vision-language-guided generation with interactive editing. A parallel line of work focuses on improving physical plausibility, PhyScene~\cite{yang2024physcene}, integrates physics-based guidance for collision avoidance and object reachability, PhysGaussian~\cite{Xie2024PhysGaussian} couples physics simulation with 3D Gaussian Splatting, and OptiScene~\cite{yang2025optiscene} applies Direct Preference Optimization to align layouts with human preferences and physical constraints.

\noindent
\textbf{VLM-Based Evaluation and Its Limitations.}
Several scene generation works~\cite{ccelen2024design, sun2025layoutvlm} adopt VLM-based evaluation, asking a model to score rendered views of generated scenes. However, VLM judges suffer from well-documented limitations: hallucinations are pervasive across large vision-language models~\cite{liu2024survey, bai2024hallucination, huang2025survey}, with studies showing that even GPT-4V fails on basic visual patterns~\cite{tong2024eyes} and that statistical biases and language priors cause systematic object hallucinations~\cite{leng2024mitigating}. IR3D-Bench~\cite{liu2025ir3d} further exposes limitations in VLM spatial precision through active 3D reconstruction. On the programmatic side, LayoutGPT~\cite{feng2023layoutgpt} reports distributional statistics (FID, KL divergence) and out-of-bound rates, PhyScene~\cite{yang2024physcene} introduces collision and reachability constraints, and frameworks like AutoLayout~\cite{chen2025autolayout} and OptiScene~\cite{yang2025optiscene} combine geometric scores with LLM ratings. The closest work to ours is SceneEval~\cite{tam2025sceneeval}, which introduces a structured evaluation framework with both fidelity metrics (object counts, attributes, spatial relationships relative to the input text) and plausibility metrics (collision, support, navigability, accessibility), grounded in SceneEval-500, a dataset of human-annotated scene descriptions. However, SceneEval still relies on an LLM for object category matching within its pipeline, its plausibility metrics are physics-oriented and do not cover semantic layout coherence such as co-occurrence plausibility, scale correctness, or canonical orientation, and it evaluates only final scenes rather than refinement trajectories. \evaluator{} addresses these gaps by grounding evaluation in dataset-derived relational priors and producing per-object, per-relationship assessments. 

\section{\dataset{} Construction}

\begin{figure}[t]
  \centering
  \includegraphics[width=1.0\textwidth]{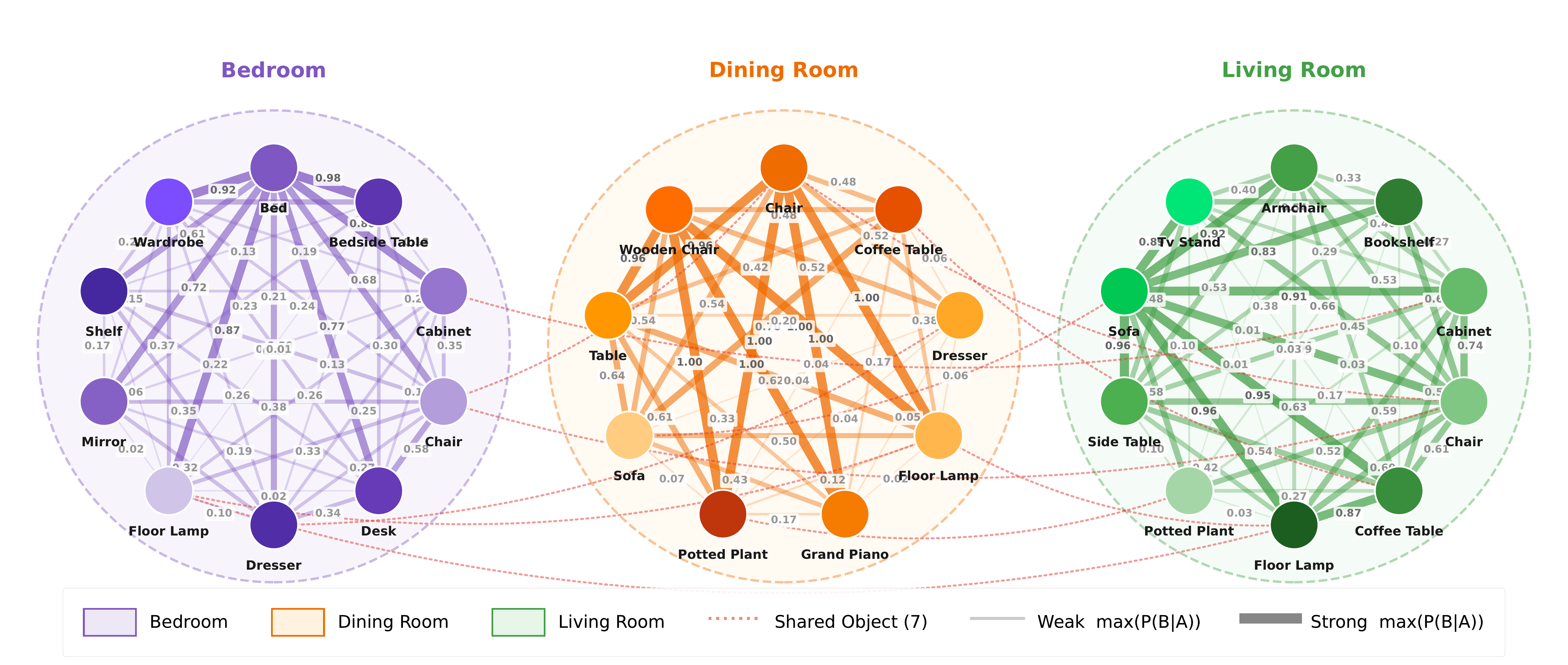}
  \vspace{-0.7cm}
  \caption{\textbf{Relational sub-graph from \dataset{} for three common room types.} Nodes represent object categories, with size proportional to frequency. Edges encode conditional co-occurrence probabilities max(P(B|A)), with thicker edges indicating stronger associations. Dotted red lines connect shared object categories across room types. \evaluator{} traverses this structure during evaluation to verify whether object co-occurrences in a generated layout are plausible for the specified room type.}
  \vspace{-0.3cm}
  \label{fig:cooccur}
\end{figure}

\evaluator{} derives its constraints from a structured relational representation of indoor scenes. We construct \dataset{} to provide this representation by aggregating spatial and semantic priors from three complementary datasets: 3D-FRONT~\cite{fu20213d}, which provides detailed geometric annotations for 6,813 professionally designed indoor scenes; ScanNet~\cite{dai2017scannet}, which contributes real-world RGB-D reconstructions of 1,511 indoor environments; and Visual Genome~\cite{krishna2017visual}, which supplies object co-occurrence and support relationships from 61,530 annotated images. From these sources, we extract four types of relational priors: object dimensions, support relations, co-occurrence statistics, and orientation relationships. The resulting ontology is organized as a per-room-type relational graph, where nodes represent object categories and edges encode conditional co-occurrence probabilities, enabling \evaluator{} to traverse object relationships during evaluation. 
\autoref{fig:cooccur} illustrates a sub-graph of \dataset{} for three common room types: bedroom, dining room, and living room.
We include full ontology details in the Appendix.

\begin{figure}[tb]
  \centering
  \includegraphics[width=1.0\textwidth]{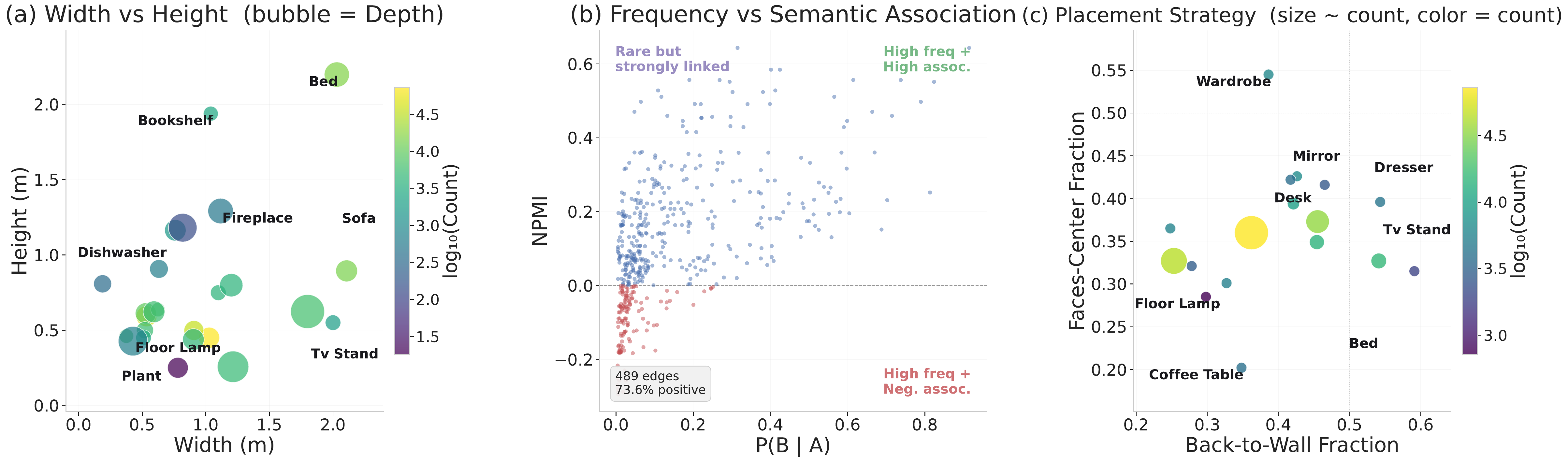}
  \vspace{-0.7cm}
  \caption{Summary statistics from \dataset{}. (Left) Object dimension distributions, with bubble size encoding depth. (Center) Co-occurrence frequency vs.\ semantic association (nPMI): 73.6\% of 489 object-pair edges show positive association. (Right) Object placement strategies, showing the fraction of instances with back-to-wall vs.\ faces-center orientation (e.g., objects like beds and coffee tables cluster toward back-to-wall, while wardrobes and mirrors favor faces-center).}
  \vspace{-0.2cm}
  \label{fig:stats}
\end{figure}

\noindent
\textbf{Dimension Extraction.} 
We extract object dimensions (width, height, and depth in meters) from 3D-FRONT and ScanNet. For 3D-FRONT, we compute bounding boxes from scene annotations with instance-specific scale parameters applied to derive metric estimates. For ScanNet, we identify per-object mesh vertices using segmentation and aggregation annotations and compute bounding boxes from the grouped vertices. For each object category, we record percentile statistics (p5, p25, median, p75, p95) along with the mean, standard deviation, and number of observations. These distributions form the geometric constraints that \evaluator{} uses for scale verification. \autoref{fig:stats} (left) visualizes the distribution of object dimensions, showing the expected scale ranges across categories.

\noindent
\textbf{Support Relations.} 
Support refers to the surface on which an object rests. We extract support information from 3D-FRONT and ScanNet through geometric reasoning: objects are processed in ascending order of their vertical position, and each is evaluated against previously placed objects whose top surface aligns with the current object's bottom within a 5~cm tolerance. Objects touching the ground are labeled as floor-supported. For Visual Genome, we retain annotated predicates corresponding to physical support interactions (e.g., \textit{on}, \textit{sitting on}, \textit{standing on}) and apply additional filtering to reduce annotation noise, including removing impossible supporting surfaces, rejecting most same-category supports, enforcing that heavier objects cannot be supported by lighter ones, and applying a spatial sanity check based on bounding box positions.

\noindent
\textbf{Co-occurrence Statistics.}
We compute pairwise object co-occurrence across all scenes, both globally and per room type, for all three datasets. For each scene, we extract the set of unique object categories present and count how many scenes contain both in the ordered pair $(a, b)$ with a certain threshold distance. We derive the conditional probability $P(b | a)$, and normalized pointwise mutual information $(n\textit{PMI})$, defined as $\textit{PMI}(a,b) / (-\log P(a,b))$, which ranges over $[-1, 1]$.
These statistics form the edges in \dataset{}'s relational graph and are what \evaluator{} uses to assess co-occurrence plausibility and proximity. \autoref{fig:stats} (center) plots co-occurrence frequency against semantic association (nPMI), revealing that 73.6\% of the 489 object-pair edges have positive association.

\noindent
\textbf{Orientation Relationships.}
We extract orientation statistics exclusively from 3D-FRONT, which provides per-object quaternion rotations and room floor geometry. We compute three types of orientation relationships. \textbf{Back-to-wall:} for categories such as sofa, bed, dresser, and bookshelf, we compute the angular difference between the object's back direction and the inward normal of the nearest wall segment. \textbf{Faces-center:} for categories such as sofa, chair, and TV, we compute the angular deviation between the object's facing direction and the direction toward the room centroid. \textbf{Faces-pair:} for each object, we compute the angular deviation between its facing direction and the direction toward every other object within a 5~m radius. The facing direction is derived from the object's yaw. For each relationship type, the ontology records the fraction of observations satisfying the constraint, along with the mean angular deviation, mean distance, and number of samples. \autoref{fig:stats} (right) shows the placement strategy landscape, with objects distributed along back-to-wall and faces-center fractions.

\section{\evaluator{} as a Symbolic Evaluator}

\evaluator{} evaluates generated layouts by traversing the relational structure encoded in \dataset{}, verifying whether the spatial arrangement of objects is consistent with the priors derived from real indoor scenes. Following the evaluation axes adopted by prior scene generation work~\cite{sun2025layoutvlm, bian2025holodeck, ran2025direct}, \evaluator{} assesses layouts along three dimensions: \textit{(a)} spatial semantics, \textit{(b)} orientation correctness, and \textit{(c)} overlap. The key difference is that where prior methods rely on VLM judges to score these criteria from rendered views, \evaluator{} resolves them symbolically by traversing \dataset{}'s constraints at the object and relationship level.

\noindent
\textbf{Spatial Semantics.}
Spatial semantics evaluates whether the objects in a scene are consistent with their expected semantic relationships. This includes whether objects are positioned in ways that align with their functional roles and typical real-world usage patterns. 
Compositionally, it evaluates whether multiple objects combine to form coherent functional groupings through their joint spatial arrangement, such as a desk and chair forming a workspace or a bed, nightstand, and lamp forming a bedside arrangement. Our spatial semantics verifier evaluates five sub-criteria: \textit{scale}, \textit{co-occurrence}, \textit{plausibility}, \textit{proximity}, and \textit{completeness}. We provide formal definitions of each in the Appendix.

\noindent
\textbf{Orientation Verification.}
The orientation verifier checks whether each object's orientation aligns with its expected placement as defined by \dataset{}. It evaluates three aspects: whether an object has its back to the nearest wall, whether it faces the room center, and whether it faces another object it is expected to be oriented toward. Each check compares the object's yaw-derived facing direction against the target direction, with the ontology specifying which checks apply to which object categories and the expected angular tolerances.

\noindent
\textbf{Overlap Detection.}
The overlap verifier checks whether objects violate basic physical constraints. It detects two types of violations: Proximity Overlap and True Overlap. Proximity Overlap detects when coarse spatial footprints of objects encroach on each other, including cases where objects are merely in close proximity without true geometric contact. True Overlap reports only cases where objects genuinely occupy shared space, filtering out false positives from loose bounding approximations. We provide implementation details in the Appendix.

\section{Experiments Setup}

\textbf{Baseline Methods.}
We test \evaluator{} on three representative 3D scene layout generation methods that span different generation strategies: LayoutGPT~\cite{feng2023layoutgpt}, which uses in-context learning to predict layouts directly; Holodeck~\cite{yang2024holodeck}, which applies constraint-based optimization over LLM-generated placements; and LayoutVLM~\cite{sun2025layoutvlm}, which jointly optimizes semantic and physical plausibility via a VLM. For each method, we use two generation backbones: Gemini-2.5-Flash and Qwen2.5-VL-72B.

\newpage
\noindent
\textbf{Baseline Evaluator.}
As the VLM-based baseline evaluator, we select Gemini-2.5-Pro, a stronger proprietary model than the generation backbones. We render the generated layouts in Blender\footnote{3D modeling and rendering package: http://www.blender.org} with Objaverse assets~\cite{deitke2023objaverse} scaled to match the object dimensions specified in the layout. From each rendered scene, we capture 2D images from multiple viewpoints (side, front, rear, and top) and provide them to the evaluator along with a prompt describing the evaluation criteria: semantic consistency, orientation correctness, and overlap.

\noindent
\textbf{Evaluation Scenes.} We evaluate on two categories of rooms. \textit{Pivotal rooms}: bedroom and living room, are the standard room types used across all baselines and prior work, enabling direct comparison. \textit{Extended rooms}: cover less common settings including bookstore, buffet restaurant, children's room, classroom, computer room, deli, dining room, game room, and florist room, which test generalization beyond the typical evaluation scope.

\section{Evaluator Reliability Analysis}
\label{sec:results}
\subsection{VLM Instability}

We first examine the reliability of VLM-based evaluation by measuring score variance across viewpoints and repeated evaluations using Gemini-2.5-Pro as the evaluator. 
\autoref{tab:per_view_variance_clean} reports the per-view variance relative to the top-view baseline across all generation methods.
Score variance is largest on the metrics where the generated scene contains the most errors. For example, LayoutVLM+Gemini produces frequent overlap errors, which explains its consistently high overlap variance across all views (11.2 Left, 28.1 Right, 17.5 Front). Additionally, left and right views represent mirrored perspectives of the same scene and should receive similar scores, yet orientation variance differs substantially between them. LayoutVLM shows high variance in the right view but low variance in the left, while other methods exhibit the opposite pattern. This asymmetry suggests the evaluator is reacting to 2D visual appearance rather than extracting 3D spatial relationships. Re-evaluating the same top-view image under identical conditions produces substantial variance: the semantic score varies by 12.46 for LayoutGPT+Gemini and 19.32 for LayoutGPT+Qwen72B. If an evaluator cannot produce consistent scores for the same input, downstream comparisons between generation methods become unreliable. We find that method rankings reverse depending on which viewpoint is chosen for evaluation.

\begin{table}[t] 
\caption{\textbf{Per-view variance relative to the top-view baseline.} Each entry is computed as 
$\frac{(\text{View} - \text{Top1})^2}{4}$ for Semantic (Sem), Orientation (Ori), and Overlap (Ovlp). The \textit{Re-eval} column measures variance when the same top-view image is evaluated twice under identical conditions. The remaining columns measure variance between the top-view and alternative viewpoints (Left, Right, Front).}
\vspace{-0.3cm}
\label{tab:per_view_variance_clean}
\centering
\setlength{\tabcolsep}{4pt}
\resizebox{\textwidth}{!}{
\begin{tabular}{@{}ll
S[table-format=2.2]
S[table-format=2.2]
S[table-format=2.2] |
S[table-format=2.2]
S[table-format=2.2]
S[table-format=2.2] |
S[table-format=2.2]
S[table-format=2.2]
S[table-format=2.2] |
S[table-format=2.2]
S[table-format=2.2]
S[table-format=2.2]@{}}
\toprule
& & \multicolumn{3}{c|}{\textbf{\textit{Re-eval}}} 
& \multicolumn{3}{c|}{\textbf{Left}} 
& \multicolumn{3}{c|}{\textbf{Right}} 
& \multicolumn{3}{c}{\textbf{Front}} \\
\cmidrule(lr){3-5}
\cmidrule(lr){6-8}
\cmidrule(lr){9-11}
\cmidrule(lr){12-14}

Method & Backbone
& {Sem.} & {Ori.} & {Ovlp}
& {Sem.} & {Ori.} & {Ovlp}
& {Sem.} & {Ori.} & {Ovlp}
& {Sem.} & {Ori.} & {Ovlp} \\
\midrule

Holodeck & Qwen72B
& 0.11 & 6.58 & 3.48
& \textbf{55.06} & \textbf{88.27} & 1.89
& \textbf{30.25} & \textbf{75.43} & 1.44
& \textbf{47.68} & \textbf{37.33} & 0.00 \\

LayoutGPT & Qwen72B
& \textbf{19.32} & 1.09 & 0.18
& 3.65 & 25.00 & 0.41
& 2.39 & 9.46 & 4.08
& 16.61 & 0.24 & \textbf{46.04} \\

LayoutVLM & Qwen72B
& 0.49 & 4.06 & 0.38
& 0.44 & 0.61 & 1.31
& 0.05 & 6.07 & 2.09
& 1.90 & 0.04 & 0.37 \\

\midrule

Holodeck & Gemini
& 0.30 & 1.97 & 0.00
& 4.69 & 8.56 & 0.00
& 0.42 & 0.09 & 7.02
& 1.22 & 1.97 & 1.56 \\

LayoutGPT & Gemini
& \textbf{12.46} & 1.90 & 1.92
& 0.12 & 35.64 & \textbf{11.36}
& 0.58 & 5.25 & 10.30
& 1.66 & 0.02 & 2.43 \\

LayoutVLM & Gemini
& 2.07 & 4.95 & \textbf{19.27}
& 4.95 & 1.39 & 11.16
& 0.18 & 9.74 & \textbf{28.09}
& 1.19 & 0.05 & 17.47 \\

\bottomrule
\end{tabular}
}
\end{table}

\begin{table}[tb]
  \caption{Comparison of Baseline methods across VLM-based evaluator and \evaluator{} on Semantic (Sem), Orientation (Ori), Overlap (Ovlp) and Average (Avg) scores. In \evaluator{}, Overlap is decomposed into ProxOvlp (Axis-Aligned Bounding Box overlap) and TrueOvlp (Oriented Bounding Box overlap).}
  \vspace{-0.3cm}
  \label{tab:baseline_heuristic_new}
  \centering
  \setlength{\tabcolsep}{3pt}
  \resizebox{\textwidth}{!}{%
  \begin{tabular}{@{}ll
    S[table-format=2.2, table-column-width=3.2em]
    S[table-format=2.2, table-column-width=3.2em]
    S[table-format=2.2, table-column-width=3.2em]
    S[table-format=2.2, table-column-width=3.2em] |
    S[table-format=2.2, table-column-width=3.2em]
    S[table-format=2.2, table-column-width=3.2em]
    S[table-format=2.2, table-column-width=3.2em]
    S[table-format=2.2, table-column-width=5.4em]
    S[table-format=2.2, table-column-width=3.2em] @{}}
    \toprule
    & & \multicolumn{4}{c|}{\textbf{VLM-Evaluation}} & \multicolumn{5}{c}{\textbf{\evaluator{}}} \\
    \cmidrule(lr){3-6} \cmidrule(lr){7-11}
    Method & Backbone & {Sem.} & {Ori.} & {Ovlp.} & {Avg.} & {Sem.} & {Ori.} & \multicolumn{2}{c}{Ovlp.} & {Avg.} \\
    \cmidrule(lr){9-10}
    & & & & & & & & {ProxOvlp} & {TrueOvlp} & \\
    \midrule
    Holodeck  & Qwen72B & 42.61 & \textbf{60.61} & \textbf{75.17} & \textbf{59.46} & 76.62 & 52.75 & \textbf{95.97} & \textbf{95.99} & 75.12 \\
    LayoutGPT & Qwen72B & \textbf{51.09} & 56.94 & 69.02 & 59.02 & 75.03 & 16.83 & 73.54 & 73.34 & 55.10 \\
    LayoutVLM & Qwen72B & 50.82 & 45.48 & 71.06 & 55.79 & \textbf{89.46} & \textbf{59.38} & 91.92 & 92.31 & \textbf{80.32} \\
    \midrule
    Holodeck  & Gemini  & 46.88 & 51.33 & \textbf{80.91} & \textbf{59.71} & 74.37 & 45.05 & \textbf{97.13} & \textbf{97.21} & 72.20 \\
    LayoutGPT & Gemini  & 48.18 & \textbf{53.82} & 77.68 & 59.89 & 74.70 & 37.52 & 76.99 & 77.79 & 63.20 \\
    LayoutVLM & Gemini  & \textbf{50.97} & 50.09 & 64.70 & 55.25 & \textbf{89.27} & \textbf{58.65} & 92.45 & 91.47 & \textbf{79.96} \\
    \bottomrule
  \end{tabular}%
  }
\end{table}

\subsection{Comparison with \evaluator{}} 

\begin{figure}[t]
  \centering
  \includegraphics[trim=1cm 5cm 1cm 5cm, clip, width=0.8\textwidth]{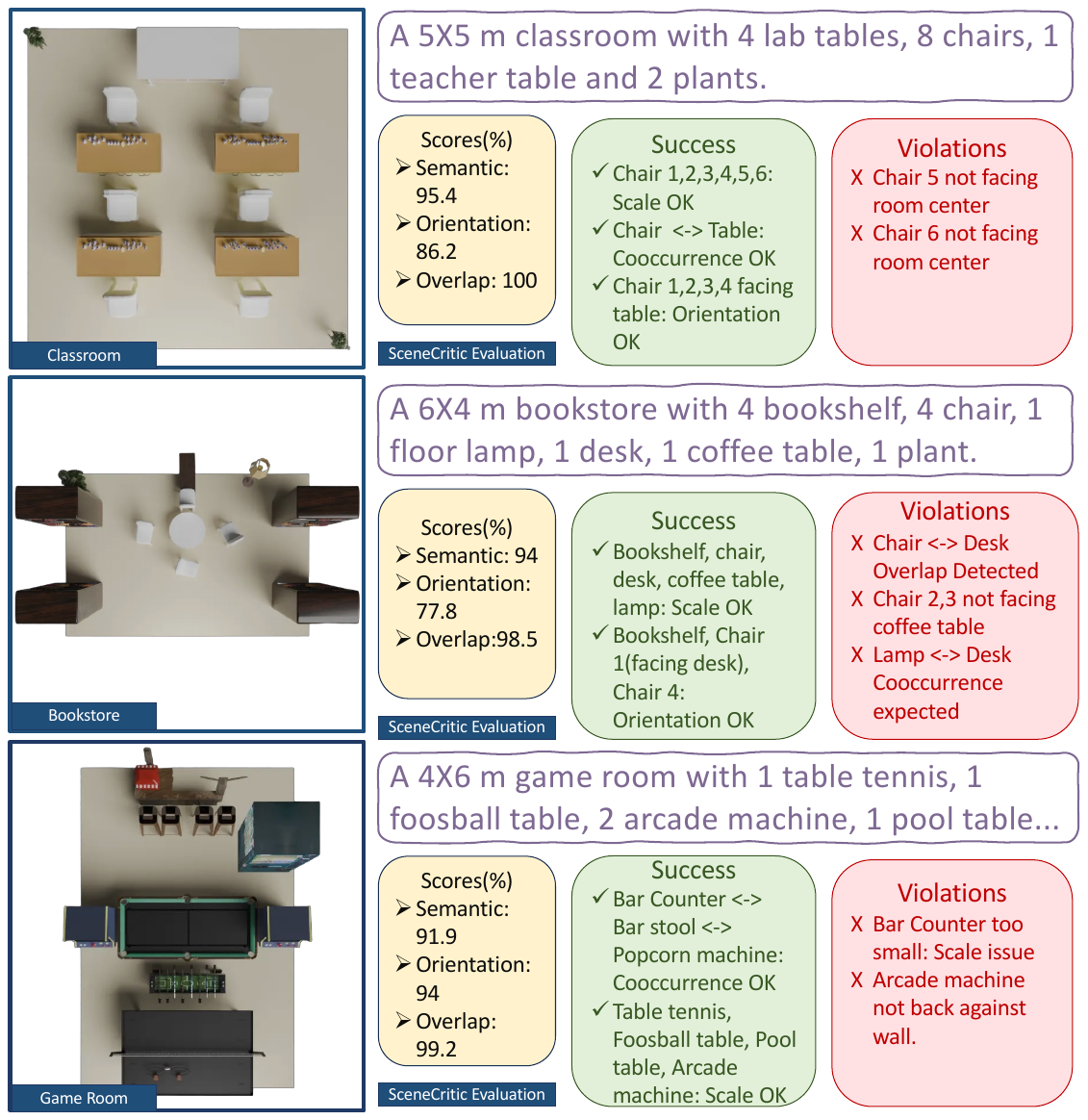}
  \caption{
  \textbf{Qualitative results of \evaluator{}}: \evaluator{} assessment for extended scenes, stating semantic, orientation and overlap scores and identifying specific violations (e.g., desk not backed against wall, cabinets not backed against wall) alongside successful placements across scale, orientation, co-occurrence, proximity, and completeness.
  }
  \vspace{-0.2cm}
  \label{fig:failure}
\end{figure}

Given these instabilities, we evaluate the same generated layouts using \evaluator{}, which operates directly on the 3D layout representation without rendered views. \autoref{tab:baseline_heuristic_new} compares the two evaluators. 
Interestingly, 
while VLM evaluation ranks Holodeck+Qwen72B highest (59.46 average), \evaluator{} ranks LayoutVLM+Qwen72B highest (80.32). 
\autoref{fig:failure} provides qualitative results; \evaluator{} produces per-object, per-constraint assessments that identify the specific violations responsible for each score, such as desks not facing nearby chairs or cabinets not backed against walls. These violations are clearly visible in the 3D layout but are inconsistently detected by the VLM evaluator from 2D renders, which explains the divergence in scores between the two evaluators.

\subsection{Human Alignment Verification}
To validate which evaluator better reflects human judgment, we conduct a pairwise evaluation study with 16 annotators providing 594 judgments across three criteria (semantic consistency, orientation correctness, and overlap) at two difficulty levels (easy and complex rooms). For reference, prior human evaluations in this space typically use 5 annotators~\cite{sun2025layoutvlm, ccelen2024design, yang2025optiscene}, while several prominent methods rely entirely on automated metrics~\cite{feng2023layoutgpt, tam2025sceneeval, xia2026sage}.

\evaluator{} achieves 94.44\% agreement with human judgments for easy scenes and 83.33\% for complex scenes. The VLM evaluator achieves 58.82\% for easy scenes and 47.06\% for complex scenes, only marginally above chance. Critically, when \evaluator{} disagrees with human judgments, the score difference is small (e.g., less than 7 points in the cases where rankings differ). When the VLM evaluator disagrees, the difference is large: Gemini assigns LayoutGPT a 75 points higher score than Holodeck in easy scenes, while human annotators rank Holodeck higher with strong inter-annotator agreement. This pattern is consistent across both difficulty levels.
We also observe that the largest VLM disagreements with humans occur on the overlap metric, whereas \evaluator{} achieves 100\% agreement with human judgments on this criterion across all comparisons. Since overlap is the most visually salient spatial error, the VLM's inability to consistently detect it from renders confirms that it is not reliably extracting geometric relationships from 2D views. We provide more details in the Appendix.

\newpage
We also compute inter-annotator agreement to assess the trustworthiness of our human evaluation study. Fleiss’ kappa values fall within the moderate range for easy rooms (0.30 for semantic, 0.46 for orientation, and 0.39 for overlap) and within the substantial range for complex rooms (0.64 for semantic, 0.59 for orientation, and 0.72 for overlap). In addition, the inter-annotator percent agreement is 81.60\% for easy rooms and 91.32\% for complex rooms, supporting the reliability of our human evaluation.

\section{Probing Spatial Reasoning through Iterative Refinement}

We design a three-stage placement pipeline that includes \textit{planning}, \textit{placement}, and \textit{refinement}, to study how models build and revise spatial structure under iterative feedback.

\subsection{Our Test Bed}

In the \textbf{planning stage}, the model receives a placement condition $\mathcal{C}_{\text{env}} = \{\mathcal{C}_{\text{desc}}, \mathcal{C}_{\text{range}}, \mathcal{C}_{\text{objects}}\}$, where $\mathcal{C}_{\text{desc}}$ is a natural language scene description, $\mathcal{C}_{\text{range}}$ defines the spatial boundaries, and $\mathcal{C}_{\text{objects}}$ lists the objects to place. The model outputs a step-by-step placement plan informed by spatial placement principles: hierarchical ordering by object size, semantic asset grouping (e.g., placing bed, nightstand, and lamp together), and navigable space preservation.

\noindent
In the \textbf{placement stage}, the model executes the plan by producing a concrete scene layout. The model is prompted with geometric constraints, including the spatial boundaries of the environment, maintaining realistic object size ratios, and assigning meaningful orientations. The output is a layout specifying each object's scale-aware bounding box, position, orientation, and category label.

\noindent
In the \textbf{refinement stage}, the model iteratively improves the layout under one of several critic modalities. We evaluate three critic types: \textit{(a)} a \textit{heuristic critic} that provides feedback based on three constraint objectives, including whether objects lie within the spatial boundary, whether all required objects are placed, and whether bounding boxes overlap; \textit{(b)} an \textit{LLM critic} that receives the layout as text and provides feedback in natural language; and \textit{(c)} a \textit{VLM critic} that receives rendered observations of the scene. We further divide VLM-based refinement into three input variants: image-only, image+text, and semantics+text. For the heuristic critic, the three objectives are combined into a single reward score with per-object feedback, allowing the model to iteratively update the layout toward a valid configuration. For LLM and VLM critics, feedback is derived from the corresponding model and the layout is modified accordingly. In all cases, \evaluator{} scores every intermediate layout, producing refinement trajectories that reveal how each model responds to criticism across successive correction steps.

\subsection{Models Evaluated}
\autoref{tab:MLLM_models} summarizes the 13 models evaluated, spanning four post-training categories: general RL (GRPO-style training), RLHF (PPO or DPO-based preference alignment), RLIF (iterative feedback RL), and RLVR (RL with verifiable rewards). Model sizes range from 7B to 235B parameters.

\begin{table*}[t]
\centering
\small
\caption{Models categorized by post-training strategy and model parameters.}
\label{tab:MLLM_models}
\resizebox{\textwidth}{!}{
\begin{tabular}{llll}
\toprule
\textbf{Model} & \textbf{Category} & \textbf{Post-Training Details} & \textbf{Params} \\
\midrule

Qwen3-14B~\cite{yang2025qwen3} & General RL & GRPO-style reinforcement learning & 14B \\
Qwen3-235B~\cite{yang2025qwen3}  & General RL & GRPO-style reinforcement learning & 235B \\
UI-Venus-Navi-72B~\cite{gu2025ui} & General RL & GRPO-based reasoning optimization & 72B \\

Gemini-2.5-flash~\cite{comanici2025gemini} & RLAIF + RLHF & SFT + Reward Model + RL & N/A \\

Qwen2.5-VL-7B-MM-RLHF~\cite{dong2024rlhf} & RLHF & PPO-style human feedback alignment & 7B \\
Qwen2.5-72B-VL~\cite{bai2025qwen25vltechnicalreport} & RLHF & SFT + DPO (preference optimization) & 72B \\
LLaMA4 Maverick~\cite{meta2024llama4} & RLHF & SFT + Online RL + DPO & $\sim$17B active (MoE) \\

Qwen3-14B-Intuitor-MATH-1EPOCH~\cite{zhao2025learning} & RLIF & Iterative feedback RL (Intuitor) & 14B \\
Qwen3-14B-GRPO-MATH-1EPOCH~\cite{zhao2025learning} & RLIF & GRPO under RLIF objective & 14B \\

Qwen2.5-14B-GRPO~\cite{shao2024deepseekmath} & RLVR & GRPO with verifiable reward & 14B \\
VL-Reasoner-72B~\cite{wang2025vl} & RLVR & GRPO + Self-Selective Revision (SSR) & 72B \\
Holo1.5-72B & RLVR & GRPO-based verifiable reward RL & 72B \\
DeepSeek-3.2V & RLVR & GRPO-style reasoning RL & 37B(active) (undisclosed) \\

\bottomrule
\end{tabular}
}
\label{tab:post_training_structured}
\end{table*}

\begin{table}[h!]
\caption{Comparison across refinement variants using MLLM backbones with \evaluator{}. Overlap is decomposed into ProxOvlp (Axis-Aligned Bounding Box overlap) and TrueOvlp (Oriented Bounding Box overlap).} 
\label{tab:pipeline_comparison}
\centering
\resizebox{0.65\textwidth}{!}{%
\begin{tabular}{@{}l ccccc@{}}
\toprule
& \multicolumn{5}{c}{\textbf{Semantic Verifier Evaluation}} \\
\cmidrule(lr){2-6}
Method & Sem. & Ori. & \multicolumn{2}{c}{Overlap} & Avg. \\
\cmidrule(lr){4-5}
& & & {ProxOvlp} & {TrueOvlp} & \\
\midrule
\multicolumn{6}{@{}l}{\textit{Heuristic}} \\
\midrule
Gemini-2.5-flash & 76.3 & 43.7 & \textbf{99.3} & \textbf{98.9} & 73.03 \\
Qwen2.5-72B-VL & 72.2 & 48.1 & 91.3 & 91.3 & 70.53 \\
DeepSeek-3.2V & \textbf{82.6} & 64.4 & 95.3 & 95.7 & \textbf{80.83} \\
LLaMA4 Maverick & 72.6 & 48.7 & 95.3 & 94.1 & 72.00 \\
Qwen3-235B & 75.0 & 47.1 & 98.2 & 97.8 & 73.37 \\
Qwen3-14B & 73.4 & 44.6 & 91.4 & 91.8 & 69.87 \\
Qwen2.5-VL-7B-MM-RLHF & 64.0 & 47.6 & 88.3 & 88.7 & 66.70 \\
Qwen2.5-14B-GRPO & 67.3 & 46.6 & 85.2 & 85.6 & 66.43 \\
Qwen3-14B-GRPO-MATH-1EPOCH & 63.8 & \textbf{65.1} & 86.2 & 87.4 & 71.90 \\
Qwen3-14B-Intuitor-MATH-1EPOCH & 65.8 & 34.3 & 87.4 & 87.9 & 62.58 \\
Holo1.5-72B & 68.3 & 48.7 & 90.0 & 90.0 & 69.00 \\
VL-Reasoner-72B & 68.9 & 42.1 & 95.6 & 95.4 & 68.83 \\
UI-Venus-Navi-72B & 66.5 & 36.6 & 91.7 & 91.7 & 64.93 \\
\midrule
\multicolumn{6}{@{}l}{\textit{LLM}} \\
\midrule
Gemini-2.5-flash & 74.2 & 50.7 & \textbf{98.3} & \textbf{98.4} & 74.42 \\
Qwen2.5-72B-VL & 72.5 & 48.6 & 89.5 & 89.3 & 70.17 \\
DeepSeek-3.2V & \textbf{78.9} & 62.8 & 97.1 & 96.9 & \textbf{79.57} \\
LLaMA4 Maverick & 69.9 & 44.7 & 89.9 & 89.5 & 68.10 \\
Qwen3-235B & 76.9 & 54.6 & 96.5 & 95.7 & 75.87 \\
Qwen3-14B & 73.3 & 49.2 & 93.4 & 93.4 & 71.97 \\
Qwen2.5-VL-7B-MM-RLHF & 65.5 & 56.9 & 90.4 & 90.6 & 70.97 \\
Qwen2.5-14B-GRPO & 67.9 & 50.7 & 86.2 & 86.6 & 68.33 \\
Qwen3-14B-GRPO-MATH-1EPOCH & 63.8 & \textbf{65.5} & 68.2 & 68.9 & 65.95 \\
Qwen3-14B-Intuitor-MATH-1EPOCH & 68.4 & 43.3 & 82.6 & 82.7 & 64.78 \\
Holo1.5-72B & 66.9 & 42.1 & 80.9 & 81.4 & 63.38 \\
VL-Reasoner-72B & 72.8 & 47.4 & 91.0 & 91.0 & 70.40 \\
UI-Venus-Navi-72B & 72.6 & 45.0 & 89.6 & 89.6 & 69.07 \\
\midrule
\multicolumn{6}{@{}l}{\textit{Image}} \\
\midrule
Gemini-2.5-flash & \textbf{78.7} & \textbf{55.8} & \textbf{97.5} & \textbf{96.6} & \textbf{77.18} \\
Qwen2.5-72B-VL & 71.2 & 49.1 & 89.0 & 88.2 & 69.63 \\
Holo1.5-72B & 68.5 & 50.2 & 84.8 & 85.1 & 67.88 \\
VL-Reasoner-72B & 71.0 & 44.0 & 87.2 & 87.2 & 67.40 \\
UI-Venus-Navi-72B & 72.2 & 49.1 & 87.7 & 87.1 & 69.57 \\
\midrule
\multicolumn{6}{@{}l}{\textit{Img+Text}} \\
\midrule
Gemini-2.5-flash & \textbf{76.7} & 48.4 & \textbf{95.7} & \textbf{96.3} & \textbf{73.70} \\
Qwen2.5-72B-VL & 71.0 & 41.8 & 89.4 & 89.0 & 67.33 \\
Holo1.5-72B & 67.9 & 49.1 & 83.6 & 84.3 & 66.98 \\
VL-Reasoner-72B & 72.4 & 44.1 & 91.1 & 90.9 & 69.17 \\
UI-Venus-Navi-72B & 72.0 & \textbf{51.8} & 90.8 & 90.6 & 71.50 \\
\midrule
\multicolumn{6}{@{}l}{\textit{Sem+Text}} \\
\midrule
Gemini-2.5-flash & \textbf{75.8} & \textbf{53.0} & \textbf{98.6} & \textbf{98.6} & \textbf{75.80} \\
Qwen2.5-72B-VL & 72.3 & 44.6 & 90.4 & 89.9 & 69.02 \\
Holo1.5-72B & 67.6 & 45.8 & 77.3 & 77.8 & 63.65 \\
VL-Reasoner-72B & 72.8 & 50.2 & 88.2 & 88.4 & 70.43 \\
UI-Venus-Navi-72B & 70.7 & 48.6 & 90.2 & 90.3 & 69.85 \\
\bottomrule
\end{tabular}%
}
\end{table}

\subsection{Results}

\textbf{Text-only LLMs can outperform VLMs on spatial reasoning:} DeepSeek-3.2V achieves the highest average scores under both heuristic refinement (80.83) and LLM-based refinement (79.57), outperforming proprietary VLMs such as Gemini-2.5-Flash. This is driven by strong semantic scores (82.6 heuristic, 78.9 LLM), suggesting that text-based spatial reasoning can be more effective than visual-semantic knowledge for layout generation.
\textbf{Orientation is the most challenging metric:} Across all models and refinement methods, orientation scores remain consistently low relative to semantic and overlap scores. Notably, Qwen2.5-14B-GRPO, a relatively small 14B model trained with a math-oriented GRPO objective, achieves the best orientation scores (65.1 heuristic, 65.5 LLM) by a significant margin, outperforming much larger models. 
We hypothesize this is because orientation verification requires precise angular computation, which may benefit from math-oriented training objectives.
\textbf{Image-based refinement is the most effective critic modality:} Using Gemini-2.5-Flash as a consistent backbone across all refinement methods, image-based VLM refinement yields the strongest performance on semantic and orientation metrics. Orientation scores increase from 43.70 (heuristic) to 50.7 (LLM) to 55.8 (image-based), an 12-point improvement. A similar pattern holds for UI-Venus-Navi-72B, whose orientation score rises from 36.6 (heuristic) to 49.1 (image-based). These results suggest that visual feedback provides richer spatial information for correcting placement and orientation than text-only inputs. Combining text with images can degrade performance, as the additional textual input may cause the model to prioritize textual cues over spatial information available in the rendered views.

\section{Conclusion}

We introduced \evaluator{}, a symbolic evaluator for indoor floor-plan layouts, and \dataset{}, a structured spatial ontology constructed from 3D-FRONT, ScanNet, and Visual Genome. Together, they provide stable, interpretable, object-level evaluation of semantic coherence, orientation correctness, and overlap, without relying on rendered views or model-based judges. 
\dataset{} is inherently extensible; by providing a simple mapping from a new dataset into our ontology format, novel room types and object categories can be readily integrated, and \evaluator{} immediately supports these additions. 
Our experiments show that \evaluator{} aligns substantially better with human judgments than VLM-based evaluators, particularly on complex scenes where VLM agreement with humans drops to 47.06\%. Through our iterative refinement testbed, we find that text-only LLMs can outperform VLMs on semantic layout quality, that math-oriented training objectives improve orientation reasoning even at small model scales, and that image-based VLM refinement is the most effective critic modality for semantic and orientation correction. 

\vspace{0.5cm}
\noindent
\textbf{Acknowledgments.}\\
This material is based upon work supported by the SUNY AI Platform on Google Cloud (Phase 1\&2). 

\newpage

\bibliographystyle{splncs04}
\bibliography{main}

\newpage
\clearpage

\appendix
\clearpage
\newpage

\section{Ontology Schema} 
We collect statistics from the diverse scene datasets and re-purpose them into our ontology data indexed by a object category. The dimension block defines object geometry constraints using percentile statistics (p5, p25, median, p75, p95, mean, std) for width, height, and depth, together with the corresponding sample counts. The room association block states how frequently a category appears in different room types, storing both the count and the fraction of occurrences relative to the total observations of that category. The support surfaces block similarly records the surfaces on which the object is observed to rest, mapping each support category (or floor) to its count and fraction. The cooccurrence block gives object co-occurrence patterns by mapping each co-occurring category to a set of statistics including the number of scenes in which both appear (count) within a threshold, the conditional probability that one will appear given the other (p\_b\_given\_a), and normalized pointwise mutual information (npmi). Co-occurrence entries are sorted by frequency and capped at 50. The cooccurrence by room block provides the same statistics conditioned on room type, with up to 50 entries per room. The orientation relationships comprises of three sub-blocks. The back to wall sub-block reports the fraction of instances whose back direction aligns with the nearest wall (within 45 degree), along with the mean angular deviation and sample count. The faces center sub-block reports the fraction of instances oriented toward the room centroid (within 60 degree). The faces pair sub-block records pairwise facing relationships, mapping other categories to the fraction of instances facing that category within 60 degree, together with the mean angle, mean distance, and number of samples. Our ontology schema has statistics for 79 unique indoor object categories.

\section{SceneCritic Setup} 
Our evaluation task is setup such that the user gives a condition to the placement generation agent $C_{env} = \{C_{desc}, C_{range}, C_{objects}\}$, where $C_{desc}$ is the natural language description of the scene to be generated, $C_{range}$ is the area range within which the objects needs to be placed and $C_{object}$ is the list of dynamic objects to be placed in the scene. Given input $C_{env}$, the placement agent outputs a scene layout composition $C^{out}_{env}$ that comprises of scale-aware bounding box ($x_i,y_i,w_i,h_i$), position $p$($x_i+w_i/2,y_i+h_i/2$), orientation ($o_i$) in degrees and object label($l_i$), where $l_i \in C_{\text{object}}$.

\noindent
\textbf{Hyperparameter Tuning.}
We ground the hyperparameter tuning in Procthor-10K dataset, a procedurally generated 3D scene dataset. We evaluate Procthor with our entire range of hyperparameters.
We measure the effectiveness of each hyperparameter by counting the number of scenes in which \evaluator{} achieves the highest score across all ProcTHOR scenes. Since ProcTHOR layouts are considered to be accurate, a hyperparameter that consistently assigns high scores to these scenes is considered appropriate.
We further constraint our choice of hyperparameter in equal cumulative percentile range from both extremes that avoids overly strict and linient choices, ensures balance between precision and recall and is robust to distribution skews( since this is based on distribution shape, not raw values which makes it invariant to scaling).

\noindent
\textbf{B.1 Scale.}
Scale Verifier acts as a key component of semantics and compositionality evaluation. A scale Verifier is expected to judge the spatial common sense of the placement agent in inferring the size of the object it has placed relative to the room size and the other placed objects, for instance, if a cupboard has a height of 8 inches, a desk placed beneath or adjacent to it should not exceed 4 inches in height. The objective of the scale verifier is to check whether each object's physical dimensions fall within the valid range defined by the ontology's 5th-to-95th percentile band. For 2D scene layouts, we compute depth using pre-computed mesh dimensions. Given mesh reference $(m_w, m_h, m_d)$ for object $l_i$ and scene bounding box $(w_i, h_i)$, scales are computed as,
\begin{equation}
    s_x = \frac{w_i}{m_w}, \qquad s_y = \frac{h_i}{m_h}, \qquad s_z = \frac{s_x + s_y}{2}, \qquad d_i = m_d \cdot s_z
\end{equation}

Given the object dimension of object $l_i$ being $dim^{l_i}_a \in \{w_i, h_i, d_i\}$ along axis $a \in \{w, h, d\}$ (width, height, depth), the per-axis verdict is:
\begin{equation}
    \mathrm{verdict}_a^{\,l_i} = \begin{cases}
        \mathit{hard\ fail} & \text{if } \mathrm{dim}_a^{\,l_i} < \dfrac{p5_a^{\,l_i}}{k} \;\;\text{or}\;\; \mathrm{dim}_a^{\,l_i} > p95_a^{\,l_i} \cdot k, \\[8pt]
        \mathit{soft\ fail} & \text{if } \mathrm{dim}_a^{\,l_i} < p5_a^{\,l_i} - \epsilon \;\;\text{or}\;\; \mathrm{dim}_a^{\,l_i} > p95_a^{\,l_i} + \epsilon, \\[4pt]
        \mathit{pass} & \text{otherwise,}
    \end{cases}
\end{equation}
where $k = 2.0$ is the hard-violation factor, $\epsilon = 0.05\;\mathrm{m}$ is the soft-violation factor.

An object passes iff all its checkable axes pass. Let $n_{\mathrm{scale}}^{\mathrm{checked}}$ denote the number of checked objects and $n_{\mathrm{scale}}^{\mathrm{passed}}$ the number that pass on every axis:
\begin{equation}
n_{scale}^{passed} = \sum_{l_i} s^{l_i}
\end{equation}

\noindent where

\begin{equation}
s^{l_i} = 
\begin{cases} 
0 & \text{if } \exists\, a, \, \text{verdict}_a^{l_i} = \text{hard fail}, \\ 
0.5 & \text{if } \forall\, a, \, \text{verdict}_a^{l_i} \neq \text{hard fail} \text{ and } \exists\, a, \, \text{verdict}_a^{l_i} = \text{soft fail}, \\ 
1 & \text{if } \forall\, a, \, \text{verdict}_a^{l_i} = \text{pass}.
\end{cases}
\end{equation}
\begin{equation}
    \boxed{S_{\mathrm{scale}} = \frac{n_{\mathrm{scale}}^{\mathrm{passed}}}{n_{\mathrm{scale}}^{\mathrm{checked}}}}
\end{equation}

\noindent
\textbf{B.2 Co-occurrence.} 
Co-occurrence Verifier evaluates Semantic Asset Grouping by verifying whether objects that commonly appear together in real-world environments are placed in close proximity within the scene. This ensures that semantically related objects maintain appropriate relational proximity and form realistic object groupings in the 3D layout. We implement co-occurrence verifier such that it assess whether pairs of objects in the scene are semantically plausible and, for strongly associated pairs, spatially proximate.

For each pair of object categories $(l_i, l_j)$ present in the scene, we compute the co-occurrence fraction:
\begin{equation}
    f_{l_i,l_j} = \max\!\bigl(P(l_i \mid l_j),\; P(l_i \mid l_j)\bigr)
\end{equation}

\noindent\textbf{Plausibility check.}\; The plausibility is evaluated as $f_{l_i,l_j} > 0.01$ for every instance pair $(l_i, l_j)$ where $l_i, l_j \in C_{obj}$.

\noindent\textbf{Proximity check.}\; Proximity check happens only if plausibility check is passed. For pairs passing plausibility test, we compute the minimum pairwise distance between two objects,
\begin{equation}
    d_{\min}(l_i, l_j) = \min_{l_i,l_j \in \mathcal C_{obj}} \|\mathbf{p}_{l_i} - \mathbf{p}_{l_j}\|_2
\end{equation}
and define the proximity indicator, evaluated once per category pair:

\begin{equation}
\mathrm{verdict}^{l_i, l_j} =
\begin{cases}
\mathit{pass} & \text{if } f_{l_i,l_j} \geq func\_thresh \;\wedge\; d_{\min}(l_i, l_j) \leq 2.0\ \mathrm{m}, \\
\mathit{pass} & \text{if } cooccur\_thresh \leq f_{l_i,l_j} < func\_thresh \;\wedge\; d_{\min}(l_i, l_j) \leq 3.5\ \mathrm{m}, \\
\mathit{fail} & \text{otherwise.}
\end{cases}
\end{equation}

where cooccurrence threshold(cooccur\_thresh) and functional threshold(func\_thresh) is set to 0.2 and 0.7 respectively.

\begin{equation}
n^{\text{passed}}_{\text{cooccur}} = \sum_{(l_i,l_j)} \mathbb{1}\!\left[\mathrm{verdict}^{l_i,l_j} = \mathit{pass}\right].
\end{equation}
\begin{equation}
    \boxed{S_{\mathrm{cooccur}} = \frac{n_{\mathrm{cooccur}}^{\mathrm{passed}}}{n_{\mathrm{cooccur}}^{\mathrm{checked}}}}
\end{equation}

\begin{table}[tb]
\centering
\caption{Distribution of threshold combinations by count for Cooccurrence.}
\label{tab:strong_functional_thresholds}
\begin{tabular}{ccc}
\toprule
\textbf{cooccur\_thresh} & \textbf{func\_thresh} & \textbf{count} \\
\midrule
0.2 & 0.5 & 678 \\
0.2 & 0.7 & 678 \\
0.5 & 0.7 & 624 \\
\bottomrule
\end{tabular}
\end{table}

While choosing the cooccurence and functional threshold values, we follow the hyperparameter tuning criteria mentioned and finalise 0.2 and 0.7 respectively as they achieve maximum scores in Procthor while not being overly linient with the evaluation. Table.~\ref{tab:strong_functional_thresholds} shows the complete distribution of score counts for cooccurrence hyperparameter combinations.

\noindent
\textbf{B.3 Completeness.}  

Completeness Verifier checks whether all the objects are placed. The objective of the completeness verifier is to measure how well the generated scene matches the expected object inventory.

For each expected object label $l_i$ with expected count $ec_{l_i}$ and actual count $ac_{l_i}$,
\begin{equation}
    \mathrm{matched}_{l_i} = \min(ac_{l_i},\, ec_{l_i}), \qquad
    \mathrm{missing}_{l_i} = (ec_{l_i} - ac_{l_i})^{+}, \qquad
    \mathrm{extra}_{l_i} = (ac_{l_i} - ec_{l_i})^{+}
\end{equation}

Each matched instance passes; each missing or extra instance fails. 

\begin{equation}
\boxed{S_{\mathrm{complete}} = \frac{\displaystyle\sum_{l_i}\, \mathrm{matched}_{l_i}}{\displaystyle\sum_{l_i}\, (e_{l_i} + \mathrm{extra}_{l_i})}}
\end{equation}

\noindent\textbf{Semantic Plausibility Score}
The weighted semantic plausibility score is:
\begin{equation}
    \boxed{S_{\mathrm{SP}}^{\mathrm{w}} = 0.33 \cdot S_{\mathrm{scale}} + 0.33 \cdot S_{\mathrm{cooccur}} + 0.33 \cdot S_{\mathrm{complete}}}
\end{equation}

\noindent
\textbf{B.4 Orientation.} The objective of Orientation Verifier is to assess whether the placed object's orientation aligns with its conventional orientation as described by \dataset{}.

The smallest angle between two orientations is:
\begin{equation}
    \Delta\theta(\alpha, \beta) = \min\!\bigl((\alpha - \beta) \bmod 360^{\circ},\;\; 360^{\circ} - (\alpha - \beta) \bmod 360^{\circ}\bigr)
\end{equation}
For object $l_i$ with yaw angle $o_i$ and target angle $\theta_{\mathrm{target}}^{\,l_i}$, the per-sub-check verdict is:
\begin{equation}
    \mathrm{verdict}_c^{\,l_i} = \begin{cases}
        \mathit{pass} & \text{if } \Delta\theta\!\bigl(o_i,\; \theta_{\mathrm{target},c}^{\,l_i}\bigr) \leq ST, \\
        \mathit{soft\ fail} & \text{if } \Delta\theta\!\bigl(o_i,\; \theta_{\mathrm{target},c}^{\,l_i}\bigr) \leq HT, \\
        \mathit{hard\ fail} & \text{otherwise,}
    \end{cases}
\end{equation}
where $c \in \mathcal{C} = \{\textit{back-to-wall},\; \textit{faces-centre},\; \textit{faces-pair}\}$ and $\theta_{\mathrm{target},c}^{\,l_i}$ is the nearest wall normal (\textit{back-to-wall}), the room centroid direction (\textit{faces-centre}), or the nearest paired-object direction (\textit{faces-pair}). ST is soft threshold of 75 degree and HT is hard threshold of 150 degree.

Each object is evaluated under multiple sub-checks, but the per-object verdict uses an \textbf{any-pass} rule: object $l_i$ receives full credit if at least one applicable sub-check passes, partial credit if at least one sub-check is a soft fail (but none pass), and zero otherwise:

\begin{equation}
s^{l_i} = 
\begin{cases} 
1 & \text{if } \exists\, c \in \mathcal{C}_{l_i} \text{ such that } \text{verdict}_c^{l_i} = \text{pass}, \\ 
0.5 & \text{if } \forall\, c \in \mathcal{C}_{l_i}, \, \text{verdict}_c^{l_i} \neq \text{pass} \text{ and } \exists\, c \in \mathcal{C}_{l_i} \text{ such that } \text{verdict}_c^{l_i} = \text{soft fail}, \\ 
0 & \text{otherwise.}
\end{cases}
\end{equation}

\noindent where $\mathcal{C}_{l_i} \subseteq \mathcal{C}$ is the set of sub-checks applicable to object $l_i$. Let $n_{orient}^{checked}$ denote the number of objects with at least one applicable sub-check and $n_{orient}^{passed}$ the accumulated score:

\begin{equation}
n_{orient}^{checked} = |\{l_i : \mathcal{C}_{l_i} \neq \varnothing\}|, \qquad n_{orient}^{passed} = \sum_{l_i : \mathcal{C}_{l_i} \neq \varnothing} s^{l_i}
\end{equation}

\begin{equation}
\boxed{S_{orient} = \frac{n_{orient}^{passed}}{n_{orient}^{checked}}}
\end{equation}

\begin{table}[tb]
\centering
\caption{Distribution of angle threshold combinations by count.}
\label{tab:angle_thresholds}
\begin{tabular}{c l c}
\toprule
\textbf{Count} & \textbf{(soft\_angle, hard\_angle) combinations} & \textbf{\# count} \\
\midrule
431 & (150, 180), (150, 270) & 2 \\
385 & (90, 150), (90, 180), (90, 270) & 3 \\
235 & (75, 90), (75, 150), (75, 180), (75, 270) & 4 \\
213 & (45, 90), (45, 150), (45, 180), (45, 270) & 4 \\
\bottomrule
\end{tabular}
\end{table}

We adopt a similar hyperparameter selection criterion to co-occurrence setting for determining the orientation thresholds. From Table.~\ref{tab:angle_thresholds}, we observe that all configurations with a soft angle threshold of 75 degree lie within the middle percentile (bottom-up $61.54\%$ and top-down $69.23\%$) band while consistently achieving a high count score, indicating stable performance across scenes. Further, for a fixed soft threshold of 75 degree, the count score remains invariant across different hard angle thresholds. This suggests that the choice of hard threshold does not significantly affect the effectiveness of the hyperparameter within this regime, and thus all such configurations are equally valid. To maintain a principled and interpretable relationship between the two thresholds, we select the hard angle threshold as twice the soft threshold, resulting in a final choice of 150 degree.

\noindent
\textbf{B.5 Overlap.}
The overlap verifier detects pairwise collisions using two complementary bounding-box representations: axis-aligned bounding boxes (AABB), which provide a fast proximity-based check, and oriented bounding boxes (OBB), which provide a tighter, rotation-aware collision test.

\paragraph{B.5.1 Proximity Overlap (AABB).}
For each object $l_i$ with center $(x_i, y_i)$, width $w_i$, and depth $h_i$, we define its 2D axis-aligned bounding box as:
\begin{equation}
    B_i^{\mathrm{AABB}} = \bigl[x_i - \tfrac{w_i}{2},\; y_i - \tfrac{h_i}{2},\; x_i + \tfrac{w_i}{2},\; y_i + \tfrac{h_i}{2}\bigr] = \bigl[x_i^{-},\; y_i^{-},\; x_i^{+},\; y_i^{+}\bigr]
\end{equation}
For each unordered pair $(l_i, l_j)$ with $i < j$, we compute the axis-wise intersection:
\begin{equation}
    \Delta x_{ij} = \min(x_i^{+},\, x_j^{+}) - \max(x_i^{-},\, x_j^{-}), \qquad
    \Delta y_{ij} = \min(y_i^{+},\, y_j^{+}) - \max(y_i^{-},\, y_j^{-})
\end{equation}
The overlap amount is:
\begin{equation}
    \mathrm{overlap}_{\mathrm{AABB}}(l_i, l_j) = \begin{cases}
        \min(\Delta x_{ij},\; \Delta y_{ij}) & \text{if } \Delta x_{ij} > 0 \;\wedge\; \Delta y_{ij} > 0, \\
        0 & \text{otherwise.}
    \end{cases}
\end{equation}
The per-pair verdict is:
\begin{equation}
    \mathrm{verdict}_{ij}^{\mathrm{AABB}} = \begin{cases}
        \mathit{fail} & \text{if } \mathrm{overlap}_{\mathrm{AABB}}(l_i, l_j) > 0.2\;\mathrm{m}\;\;(\text{tolerance}), \\
        \mathit{pass} & \text{otherwise.}
    \end{cases}
\end{equation}
Let $n_{\mathrm{AABB}}^{\mathrm{checked}} = \binom{N}{2}$ denote the total number of pairs evaluated and $n_{\mathrm{AABB}}^{\mathrm{passed}}$ the number that pass:
\begin{equation}
    n_{\mathrm{AABB}}^{\mathrm{passed}} = \bigl|\bigl\{(i,j) : \mathrm{verdict}_{ij}^{\mathrm{AABB}} = \mathit{pass}\bigr\}\bigr|
\end{equation}
\begin{equation}
    \boxed{S_{\mathrm{ProxOvlp}} = \frac{n_{\mathrm{AABB}}^{\mathrm{passed}}}{n_{\mathrm{AABB}}^{\mathrm{checked}}}}
\end{equation}

\paragraph{B.5.2 True Overlap (OBB).}
For each object $l_i$ with center $(x_i, y_i)$, width $w_i$, depth $h_i$, and yaw angle $o_i$, we define its oriented bounding box as the set of four corner vertices:
\begin{equation}
    B_i^{\mathrm{OBB}} = \left\{ \begin{pmatrix} x_i \\ y_i \end{pmatrix} + R(o_i) \begin{pmatrix} \pm\, w_i/2 \\ \pm\, h_i/2 \end{pmatrix} \right\}, \qquad
    R(\theta) = \begin{pmatrix} \cos\theta & -\sin\theta \\ \sin\theta & \cos\theta \end{pmatrix}
\end{equation}
where $R(o_i)$ is the 2D rotation matrix. Each $B_i^{\mathrm{OBB}}$ is a convex polygon. For each unordered pair $(l_i, l_j)$ with $i < j$, we test for intersection using the \textit{Separating Axis Theorem} (SAT). The candidate separating axes are the four edge normals of the two OBBs:
\begin{equation}
    \mathcal{A}_{ij} = \bigl\{\mathbf{n}_1^{(i)},\; \mathbf{n}_2^{(i)},\; \mathbf{n}_1^{(j)},\; \mathbf{n}_2^{(j)}\bigr\}
\end{equation}
where $\mathbf{n}_1^{(i)}, \mathbf{n}_2^{(i)}$ are the two unique edge normals of $B_i^{\mathrm{OBB}}$. For each axis $\mathbf{a} \in \mathcal{A}_{ij}$, we project both OBBs onto $\mathbf{a}$, obtaining the scalar intervals $[\min_k(\mathbf{v}_k^{(i)} \cdot \mathbf{a}),\; \max_k(\mathbf{v}_k^{(i)} \cdot \mathbf{a})]$ and $[\min_k(\mathbf{v}_k^{(j)} \cdot \mathbf{a}),\; \max_k(\mathbf{v}_k^{(j)} \cdot \mathbf{a})]$, and compute their overlap:
\begin{equation}
    d(\mathbf{a}) = \min\!\bigl(\max_k(\mathbf{v}_k^{(i)} \cdot \mathbf{a}),\; \max_k(\mathbf{v}_k^{(j)} \cdot \mathbf{a})\bigr) - \max\!\bigl(\min_k(\mathbf{v}_k^{(i)} \cdot \mathbf{a}),\; \min_k(\mathbf{v}_k^{(j)} \cdot \mathbf{a})\bigr)
\end{equation}
where $\mathbf{v}_k^{(i)}$ denotes the $k$-th vertex of $B_i^{\mathrm{OBB}}$. If a separating axis exists (i.e.\ $d(\mathbf{a}) < 0$ for some $\mathbf{a}$), the OBBs do not intersect. The penetration depth is:
\begin{equation}
    \mathrm{overlap}_{\mathrm{OBB}}(l_i, l_j) = \begin{cases}
        \displaystyle\min_{\mathbf{a} \in \mathcal{A}_{ij}} d(\mathbf{a}) & \text{if } d(\mathbf{a}) > 0 \;\;\forall\; \mathbf{a} \in \mathcal{A}_{ij}, \\
        0 & \text{otherwise.}
    \end{cases}
\end{equation}
The per-pair verdict applies the same tolerance:
\begin{equation}
    \mathrm{verdict}_{ij}^{\mathrm{OBB}} = \begin{cases}
        \mathit{fail} & \text{if } \mathrm{overlap}_{\mathrm{OBB}}(l_i, l_j) > 0.2\;\mathrm{m}\;\;(\text{tolerance}), \\
        \mathit{pass} & \text{otherwise.}
    \end{cases}
\end{equation}
\begin{equation}
    n_{\mathrm{OBB}}^{\mathrm{passed}} = \bigl|\bigl\{(i,j) : \mathrm{verdict}_{ij}^{\mathrm{OBB}} = \mathit{pass}\bigr\}\bigr|
\end{equation}
\begin{equation}
    \boxed{S_{\mathrm{TrueOvlp}} = \frac{n_{\mathrm{OBB}}^{\mathrm{passed}}}{n_{\mathrm{OBB}}^{\mathrm{checked}}}}
\end{equation}

\paragraph{B.5.3 Combined Overlap Score.}
The final overlap score used in computing the overall average is the mean of the two measures:
\begin{equation}
    \boxed{S_{\mathrm{overlap}} = \frac{S_{\mathrm{ProxOvlp}} + S_{\mathrm{TrueOvlp}}}{2}}
\end{equation}

\begin{figure}[H]
  \centering
  \includegraphics[trim=0cm 2.5cm 0cm 2.5cm, clip, width=1.0\textwidth]{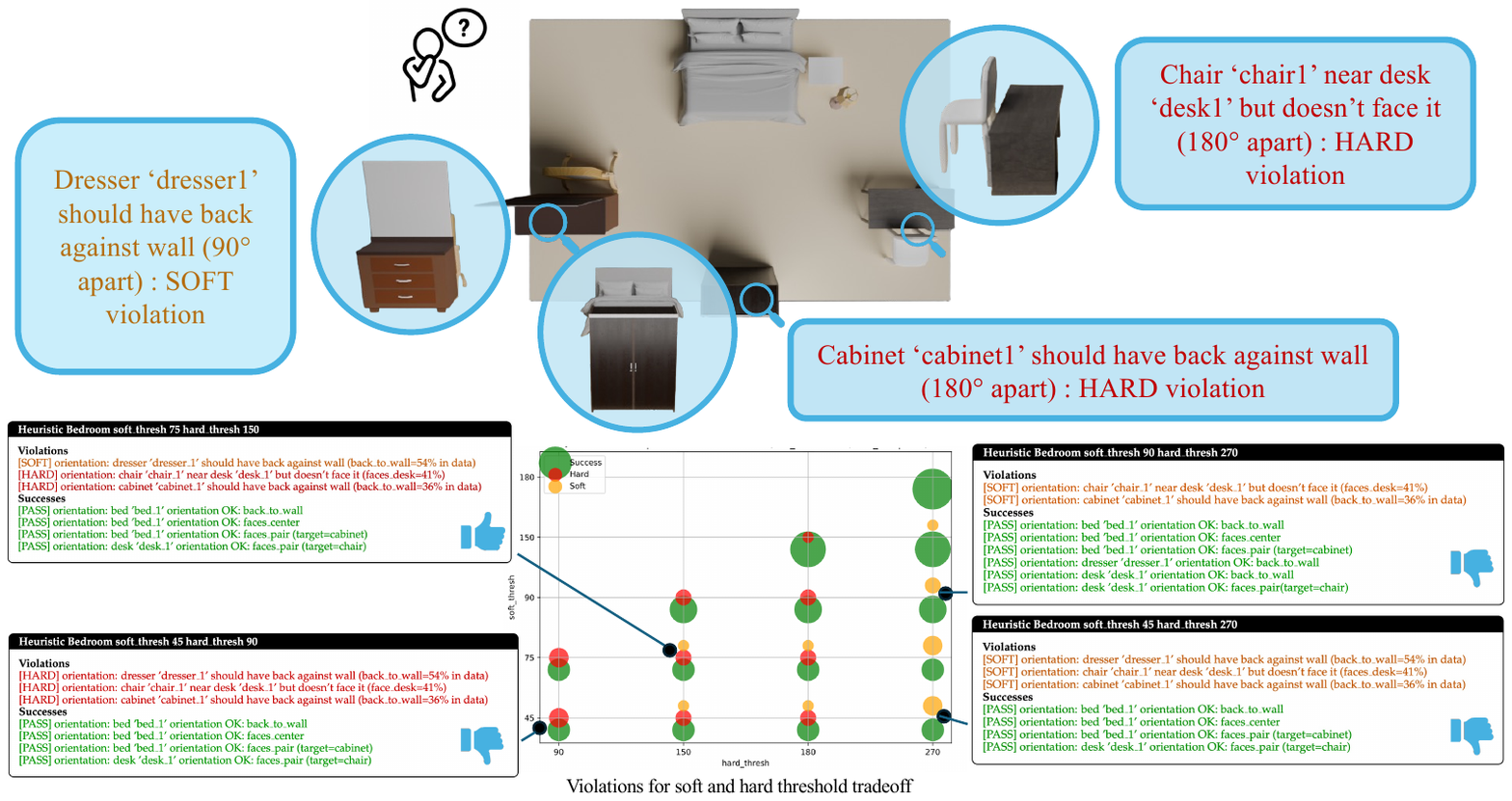}
  \caption{\textbf{Impact of hyperparameter choice on \evaluator{} Evaluation:} \evaluator{} shows strong alignment with human judgment under the selected hyperparameter configuration. In contrast, alternative hyperparameter settings lead to systematic biases, causing the evaluator to behave either overly lenient or excessively strict.
  }
  \label{ablation}
\end{figure}

\section{Additional Details}
\noindent
\textbf{Analysing hyperparameter choice on our testbed generation:}
Fig.~\ref{ablation} shows how accurately our Procthor grounded method of hyperparameter choice aligns with human judgement for our test bed generated scenes. We evaluate a typical bedroom generated by heuristic refinement method and show the success, violations detected by SceneCritic for some significant orientation hyperparameter combinations. Additionally, the plot visualizes the distribution of success, hard failures, and soft failures for all possible hyperparameter combinations where the size of each blob is proportional to its corresponding count. We find the analysis generated by our hyperparameter choice aligns with human judgement that also validates our hyperparameter tuning process grounded on Procthor-10K.

\noindent
\textbf{Human evaluation details.} 
Table.~\ref{tab:human_study_easy_clean} and Table.~\ref{tab:human_study_complex_clean} summarises the exact human study conducted for validating \evaluator{}.

\begin{table*}[t]
\caption{Comparison across all pipeline variants evaluated across multiple rendered views.
Best results per column within each view are in \textbf{bold}.}
\label{tab:all_views_eval_comparison}
\centering
\small
\resizebox{\textwidth}{!}{%
\begin{tabular}{@{}l
c c c >{\columncolor{gray!15}}c
c c c >{\columncolor{gray!15}}c
c c c >{\columncolor{gray!15}}c
c c c >{\columncolor{gray!15}}c
c c c >{\columncolor{gray!15}}c@{}}
\toprule
& \multicolumn{4}{c}{\textbf{Top-view (1)}} 
& \multicolumn{4}{c}{\textbf{Top-view (2)}} 
& \multicolumn{4}{c}{\textbf{Left view}} 
& \multicolumn{4}{c}{\textbf{Right view}}
& \multicolumn{4}{c}{\textbf{Front view}} \\
\cmidrule(lr){2-5} 
\cmidrule(lr){6-9} 
\cmidrule(lr){10-13} 
\cmidrule(lr){14-17}
\cmidrule(lr){18-21}
Method 
& Sem. & Ori. & Ovlp. & Avg.
& Sem. & Ori. & Ovlp. & Avg.
& Sem. & Ori. & Ovlp. & Avg.
& Sem. & Ori. & Ovlp. & Avg.
& Sem. & Ori. & Ovlp. & Avg. \\
\midrule
\multicolumn{21}{@{}l}{\textit{Baselines}} \\
\midrule

Holodeck + Gemini 
& 46.88 & 51.33 & 80.91 & 59.71
& 47.97 & 48.52 & 80.83 & 59.11
& 51.21 & 45.48 & 80.92 & 59.21
& 48.18 & 50.73 & 75.61 & 58.17
& 49.09 & 48.52 & 78.41 & 58.67 \\

Holodeck + Qwen72B
& 42.61 & 60.61 & 75.17 & 59.46
& 41.94 & 55.48 & 71.44 & 56.29
& 57.45 & 41.82 & 72.42 & 57.23
& 53.61 & 43.24 & 72.77 & 56.54
& 56.42 & 48.39 & 75.17 & 59.99 \\

LayoutGPT + Gemini
& 48.18 & 53.82 & 77.68 & 59.89
& 55.24 & 56.58 & 74.91 & 62.24
& 47.48 & 41.88 & 70.94 & 53.43
& 49.70 & 49.24 & 71.26 & 56.73
& 50.76 & 54.12 & 74.56 & 59.81 \\

LayoutGPT + Qwen72B
& 51.09 & 56.94 & 69.02 & 59.02
& 59.88 & 59.03 & 68.18 & 62.36
& 47.27 & 46.94 & 70.30 & 54.84
& 54.18 & 50.79 & 73.06 & 59.34
& 59.24 & 55.97 & 55.45 & 56.89 \\

LayoutVLM + Gemini
& 50.97 & 50.09 & 64.70 & 55.25
& 48.09 & 45.64 & 73.48 & 55.74
& 55.42 & 47.73 & 71.38 & 58.18
& 50.12 & 43.85 & 75.30 & 56.42
& 53.15 & 50.52 & 73.06 & 58.91 \\

LayoutVLM + Qwen72B
& 50.82 & 45.48 & 71.06 & 55.79
& 49.42 & 41.45 & 72.29 & 54.39
& 52.15 & 43.91 & 73.35 & 56.47
& 51.24 & 40.55 & 73.95 & 55.25
& 53.58 & 45.09 & 72.27 & 56.98 \\

\bottomrule
\end{tabular}}
\end{table*}

\begin{table*}[t]
\centering
\caption{Pairwise comparison — Difficulty: Easy. 
Per-row inter-annotator agreement (IAA) is computed as $\max(\text{A Wins},\text{B Wins})/\text{Total}$. 
We additionally report row-wise Fleiss' $\kappa$ using global label marginals for each metric across all Easy comparisons.
\evaluator{} Agree = 94.44\% (IAA); Aggregated IAA = 77.08\% / 83.33\% / 84.38\% (Sem/Ori/Ovlp); Fleiss' $\kappa$ = 0.30 / 0.46 / 0.39; VLM Agree = 58.82\%.} 
\label{tab:human_study_easy_clean}
\resizebox{\textwidth}{!}{
\begin{tabular}{ll|cc|cc|ccccc|ccccc|ccccc|ccc|ccc|ccc|ccc}
\toprule
\multirow{2}{*}{Method A} & \multirow{2}{*}{Method B}
& \multicolumn{2}{c|}{Method A Scores}
& \multicolumn{2}{c|}{Method B Scores}
& \multicolumn{5}{c|}{\textbf{Semantic}}
& \multicolumn{5}{c|}{\textbf{Orientation}}
& \multicolumn{5}{c|}{\textbf{Overlap}}
& \multicolumn{3}{c|}{\textbf{IAA (\%)}}
& \multicolumn{3}{c|}{\textbf{Fleiss' $\kappa$}}
& \multicolumn{3}{c|}{\textbf{VLM Agree}}
& \multicolumn{3}{c}{\textbf{\evaluator{} Agree}} \\
\cmidrule(lr){3-4}\cmidrule(lr){5-6}
\cmidrule(lr){7-11}
\cmidrule(lr){12-16}
\cmidrule(lr){17-21}
\cmidrule(lr){22-24}
\cmidrule(lr){25-27}
\cmidrule(lr){28-30}
\cmidrule(lr){31-33}

& & \evaluator{} & VLM & \evaluator{} & VLM
& A W & B W & Tot & A\% & B\%
& A W & B W & Tot & A\% & B\%
& A W & B W & Tot & A\% & B\%
& Sem & Ori & Ovlp
& Sem & Ori & Ovlp
& Sem & Ori & Ovlp
& Sem & Ori & Ovlp \\
\midrule

heuristic & holodeck
& 75.6 & 60 & 71.3 & 25
& 9 & 7 & 16 & 56.2 & 43.7
& 12 & 4 & 16 & 75.0 & 25.0
& 11 & 5 & 16 & 68.8 & 31.2
& 62.5 & 75.0 & 68.8
& -0.098 & 0.126 & -0.134
& Y & Y & N
& Y & Y & Y \\

heuristic & layoutgpt
& 75.6 & 60 & 53.6 & 100
& 15 & 1 & 16 & 93.8 & 6.2
& 14 & 2 & 16 & 87.5 & 12.5
& 16 & 0 & 16 & 100.0 & 0.0
& 93.8 & 87.5 & 100.0
& 0.739 & 0.490 & 1.000
& N & Y & -
& Y & Y & Y \\

heuristic & layoutvlm
& 75.6 & 60 & 81.9 & 60
& 12 & 4 & 16 & 75.0 & 25.0
& 14 & 2 & 16 & 87.5 & 12.5
& 11 & 5 & 16 & 68.8 & 31.2
& 75.0 & 87.5 & 68.8
& 0.164 & 0.490 & -0.134
& - & Y & N
& N & Y & Y \\

holodeck & layoutgpt
& 71.3 & 25 & 53.6 & 100
& 14 & 2 & 16 & 87.5 & 12.5
& 15 & 1 & 16 & 93.8 & 6.2
& 15 & 1 & 16 & 93.8 & 6.2
& 87.5 & 93.8 & 93.8
& 0.512 & 0.727 & 0.691
& N & N & Y
& Y & Y & Y \\

holodeck & layoutvlm
& 71.3 & 25 & 81.9 & 60
& 6 & 10 & 16 & 37.5 & 62.5
& 6 & 10 & 16 & 37.5 & 62.5
& 14 & 2 & 16 & 87.5 & 12.5
& 62.5 & 62.5 & 87.5
& -0.045 & -0.093 & 0.423
& Y & Y & N
& Y & Y & Y \\

layoutgpt & layoutvlm
& 53.6 & 100 & 81.9 & 60
& 2 & 14 & 16 & 12.5 & 87.5
& 1 & 15 & 16 & 6.2 & 93.8
& 2 & 14 & 16 & 12.5 & 87.5
& 87.5 & 93.8 & 87.5
& 0.512 & 0.727 & 0.423
& N & - & Y
& Y & Y & Y \\

\bottomrule
\end{tabular}}
\end{table*}

\begin{table*}[h!]
\centering
\caption{Pairwise comparison — Difficulty: Complex. 
Per-row inter-evaluator agreement (IAA) is computed as $\max(\text{A Wins},\text{B Wins})/\text{Total}$. 
We additionally report row-wise Fleiss' $\kappa$ using global label marginals.
\evaluator{} Agree = 83.33\% (IAA); Aggregated IAA = 90.63\% / 89.58\% / 93.75\% (Sem/Ori/Ovlp); Fleiss' $\kappa$ = 0.64 / 0.59 / 0.72; VLM Agree = 47.06\%.} 
\label{tab:human_study_complex_clean}
\resizebox{\textwidth}{!}{
\begin{tabular}{ll|cc|cc|ccccc|ccccc|ccccc|ccc|ccc|ccc|ccc}
\toprule
\multirow{2}{*}{Method A} & \multirow{2}{*}{Method B}
& \multicolumn{2}{c|}{Method A Scores}
& \multicolumn{2}{c|}{Method B Scores}
& \multicolumn{5}{c|}{\textbf{Semantic}}
& \multicolumn{5}{c|}{\textbf{Orientation}}
& \multicolumn{5}{c|}{\textbf{Overlap}}
& \multicolumn{3}{c|}{\textbf{IAA (\%)}}
& \multicolumn{3}{c|}{\textbf{Fleiss' $\kappa$}}
& \multicolumn{3}{c|}{\textbf{VLM Agree}}
& \multicolumn{3}{c}{\textbf{\evaluator{} Agree}} \\
\cmidrule(lr){3-4}\cmidrule(lr){5-6}
\cmidrule(lr){7-11}
\cmidrule(lr){12-16}
\cmidrule(lr){17-21}
\cmidrule(lr){22-24}
\cmidrule(lr){25-27}
\cmidrule(lr){28-30}
\cmidrule(lr){31-33}

& & \evaluator{} & VLM & \evaluator{} & VLM
& A W & B W & Tot & A\% & B\%
& A W & B W & Tot & A\% & B\%
& A W & B W & Tot & A\% & B\%
& Sem & Ori & Ovlp
& $\kappa_S$ & $\kappa_O$ & $\kappa_L$
& Sem & Ori & Ovlp
& Sem & Ori & Ovlp \\
\midrule

heuristic & holodeck
& 79.39 & 75 & 79.5 & 95
& 13 & 3 & 16 & 81.2 & 18.8
& 13 & 3 & 16 & 81.2 & 18.8
& 14 & 2 & 16 & 87.5 & 12.5
& 81.2 & 81.2 & 87.5
& 0.426 & 0.383 & 0.563
& N & N & N
& N & N & Y \\

heuristic & layoutgpt
& 79.39 & 75 & 68.3 & 100
& 14 & 2 & 16 & 87.5 & 12.5
& 14 & 2 & 16 & 87.5 & 12.5
& 16 & 0 & 16 & 100.0 & 0.0
& 87.5 & 87.5 & 100.0
& 0.562 & 0.530 & 1.000
& N & Y & N
& Y & Y & Y \\

heuristic & layoutvlm
& 79.39 & 75 & 94.0 & 95
& 3 & 13 & 16 & 18.8 & 81.2
& 1 & 15 & 16 & 6.2 & 93.8
& 2 & 14 & 16 & 12.5 & 87.5
& 81.2 & 93.8 & 87.5
& 0.426 & 0.787 & 0.563
& Y & Y & Y
& Y & Y & N \\

holodeck & layoutgpt
& 79.5 & 95 & 68.3 & 100
& 16 & 0 & 16 & 100.0 & 0.0
& 14 & 2 & 16 & 87.5 & 12.5
& 15 & 1 & 16 & 93.8 & 6.2
& 100.0 & 87.5 & 93.8
& 1.000 & 0.530 & 0.770
& N & Y & Y
& Y & Y & Y \\

holodeck & layoutvlm
& 79.5 & 95 & 94.0 & 95
& 1 & 15 & 16 & 6.2 & 93.8
& 1 & 15 & 16 & 6.2 & 93.8
& 1 & 15 & 16 & 6.2 & 93.8
& 93.8 & 93.8 & 93.8
& 0.787 & 0.787 & 0.770
& - & N & N
& Y & Y & Y \\

layoutgpt & layoutvlm
& 68.3 & 100 & 94.0 & 95
& 0 & 16 & 16 & 0.0 & 100.0
& 1 & 15 & 16 & 6.2 & 93.8
& 0 & 16 & 16 & 0.0 & 100.0
& 100.0 & 93.8 & 100.0
& 1.000 & 0.787 & 1.000
& N & Y & Y
& Y & Y & Y \\

\bottomrule
\end{tabular}}
\end{table*}

\end{document}